\documentclass{article}

\usepackage{microtype}
\usepackage{graphicx}
\usepackage{subcaption}
\usepackage{booktabs}
\usepackage{multirow}
\usepackage{tabularx}
\usepackage[table]{xcolor}
\usepackage{colortbl}
\usepackage{stfloats}

\usepackage{enumitem}
\setlist{
    itemsep=2pt,      
    topsep=2pt,       
    parsep=1pt,       
    partopsep=0pt     
}

\usepackage{hyperref}

\usepackage[accepted]{icml2026}

\usepackage{amsmath}
\usepackage{amssymb}
\usepackage{mathtools}
\usepackage{amsthm}

\usepackage[capitalize,noabbrev]{cleveref}

\theoremstyle{plain}

\theoremstyle{definition}

\theoremstyle{remark}

\usepackage[textsize=tiny]{todonotes}

\icmltitlerunning{Hard or Just Unreached? Diagnosing the Sampling Blind Spot in Math-Reasoning Difficulty Estimation}

\begin{document}

\twocolumn[
\icmltitle{Hard or Just Unreached? \\Diagnosing the Sampling Blind Spot in Math-Reasoning Difficulty Estimation}

  \icmlsetsymbol{equal}{*}

  \begin{icmlauthorlist}
    \icmlauthor{Luca Zhou}{sapienza}
    \icmlauthor{Sajel Shah}{notdiamond}
    \icmlauthor{Emanuele Rodolà}{sapienza,paradigma}
    \icmlauthor{Roberto Dessì}{notdiamond}
  \end{icmlauthorlist}

  \icmlaffiliation{sapienza}{Sapienza University of Rome}
  \icmlaffiliation{notdiamond}{Not Diamond}
  \icmlaffiliation{paradigma}{Paradigma}

  \icmlcorrespondingauthor{Luca Zhou}{luca.zhou@uniroma1.it}

  \icmlkeywords{Machine Learning, LLMs}

  \vskip 0.3in
]



\printAffiliationsAndNotice{}  

\begin{abstract}
Math and science reasoning benchmarks rely on pass@$k$, the
fraction of sampled chains that reach gold, as the canonical
per-example difficulty signal. The same signal drives RL with
verifiable rewards, math data curation, synthetic curricula,
and verifier training. We show this proxy has a persistent blind spot on its hardest stratum: on the eight free-form math cells we test (GSM8K and MATH across four open-weight models), $10.3$--$22.9\%$ of the examples that no sampling seed solves in six tries are instead solved at matched compute by a six-chain deterministic regime. These are greedy decoding plus five cheap residual-stream perturbations applied via activation grafting, while greedy alone solves at most $6\%$ on these math cells.
Recovery scales with the additional budget,
across perturbations whose mechanistic distinctness we verify
across all twelve cells (cross-kind fix-set Jaccard $\leq\!0.47$
in every setup. Activation grafting is used as an intervention on internal representations, not a
decoding method; we use it purely as a diagnostic and diversification
tool, and our recovered items show that the pass@$k\!=\!0$ stratum is structurally identifiable in the residual stream rather than that the unmodified model reaches them under ordinary inference.
\end{abstract}

\begin{figure}[t]
\centering
\includegraphics[width=0.9\linewidth]{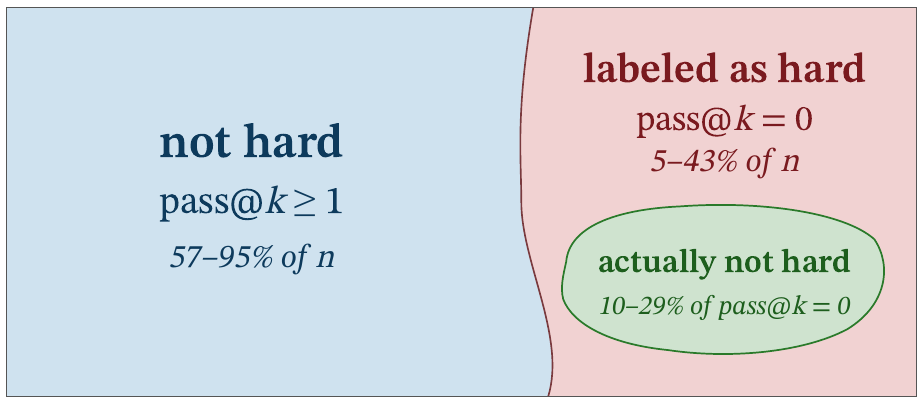}
  \caption{\textit{The sampling blind spot (n=$1000$).}
    Across twelve (model, benchmark) cells,
    pass@$k$ flags $5\text{--}43\%$ of items as ``hard'' (red, $\text{pass@}k\!=\!0$).
    A compute-matched deterministic regime (six greedy chains with residual-stream interventions) reaches $10\text{--}29\%$ of that stratum (green), showing that a non-trivial fraction of items pass@$k$ calls ``hard'' are structurally identifiable in the model's residual stream rather than intrinsically hard.}
\label{fig:teaser}
\end{figure}

\section{Introduction}

RL pipelines with verifiable rewards, math/code data curation,
difficulty-stratified curricula, and pass@$k$-style hardness
annotations all share a single label-required primitive: examples
on which no sampled chain reaches gold are filtered out, downweighted,
or labelled
hard~\citep{lambert2024tulu3,deepseekai2025r1,shao2024deepseekmath,yuan2023rft,toshniwal2024openmathinstruct,zhou2025cats}.
The signal is the pass@$k$ count itself, the number of sampled
chains that reach gold~\citep{cobbe2021gsm8k,chen2021evaluating};
where a single answer is needed, $k$ chains are aggregated by
majority vote or
self-evaluation~\citep{wang2022selfconsistency,chen2023universal,aggarwal2023letscalmly}.
The first sample, the second, the $k$th are all drawn the same way,
and temperature sampling is treated as \emph{the} source of decoding
diversity. A problem that no sampling chain solves is treated, by some
reasonable proxy, as difficult.

We show this last assumption fails on a structurally identifiable
slice. Through three empirical observations across four open-weight
instruction-tuned models (\texttt{Qwen-2.5-3B},
\texttt{Llama-3.2-3B}, \texttt{Llama-3.1-8B},
\texttt{Mistral-Nemo-12B}) and three reasoning
benchmarks (GSM8K, MATH, MMLU-Pro), we argue that sampling has a
\emph{persistent blind spot} on its hardest stratum at the budgets
we test, and that a non-trivial fraction of the slice flagged as
hardest by sampling-only estimators is in fact reachable by
deterministic decoding at matched compute.

\textbf{(1) Greedy is competitive at a single sample.}
Across the (model, benchmark) cells we test, greedy decoding is
competitive with or better than the mean single-sample accuracy,
with the gap reaching up to $2.8\%$. The default single-pass
baseline for self-consistency is therefore already weaker than a
deterministic alternative on the same compute.

\textbf{(2) Sampling has a persistent blind spot on its hardest
stratum.}
The pass@$k\!=\!0$ stratum, examples no sample reaches gold in
$k$ tries, is the operative signal for any pipeline whose
deployment already has labels: RL with verifiable rewards,
math data curation, and pass@$k$-style hardness annotation, etc. At
$k\!=\!6$ this stratum is large on free-form math: $5.1$--$8.3\%$
of GSM8K and $28.7$--$43.5\%$ of MATH across the four models we
test ($51$--$435$ examples per setup; Table~\ref{tab:Rk}). The
per-sample marginal-shrink rate decreases monotonically through our
sampling budget (Fig.~\ref{fig:shrink}; App.~\ref{app:robustness}),
so the stratum is not an artifact of an undersized $k$.
This stratum is treated, by some reasonable proxy, as the slice the model cannot solve. The
question we focus on: is that slice intrinsically hard, or is
it reachable but only off the stochastic axis? See the intuition in Fig.~\ref{fig:teaser}.

\textbf{(3) The deterministic regime reaches what six samples miss.}
We isolate the slice on which sampling truly fails, examples no
sampling seed reaches in six tries, and ask how much of it a
deterministic regime at matched compute reaches. Activation
grafting is the diagnostic: replacing the last prompt-token hidden
state with cheap synthetic vectors, then decoding greedily, holds
prompt and weights fixed while perturbing only the deterministic
trajectory. A six-chain deterministic regime (greedy plus five
grafts) reaches $10.3$--$22.9\%$ of these examples on the
eight free-form math cells (GSM8K, MATH; $10$--$29\%$ across
all twelve cells), while greedy alone reaches only up to $9\%$
(Sec.~\ref{sec:diversify}). The recovery
scales with deterministic budget across five graft kinds whose
mechanistic distinctness, we verify across all twelve cells
(Sec.~\ref{sec:diversify}, App.~\ref{app:extragraft}: cross-kind
fix-set Jaccard $\leq\!0.47$ in every cell, with the single
same-axis pair of random probes reaching $0.76$ on
\texttt{Qwen-2.5-3B} / GSM8K as an illustrative ceiling). This rules out both the ``intrinsically hard'' and the ``one privileged vector'' readings. The point is
not that deterministic beats sampling on aggregate accuracy; it is
that the two regimes have partially disjoint correct sets, and the slice flagged as hardest by sampling-only proxies might not be the actual hardest slice for the model.

\textbf{The logical chain.}
Sampling has a persistent blind spot on its hardest stratum, and
the deterministic axis reaches into it: recovery on the
no-sample-reaches-gold stratum scales with the deterministic
budget (one chain $\to$ six chains: $0$--$9\%\!\to\!10$--$29\%$)
across mechanistically distinct grafts, so what looked hard under sampling was, for a non-trivial fraction, just unreached under stochastic decoding.

\textbf{Implication.}
On the (model, benchmark) cells we test, sampling-derived
difficulty estimates (pass@$k$ filters, RL data-curation
pipelines, hardness annotations, difficulty-stratified curricula)
mislabel a structurally identifiable fraction of their hardest
stratum: $10$--$29\%$ of the no-sample-reaches-gold slice is
reached by a six-chain deterministic regime at matched compute. Hence, difficulty estimates
derived from sampling alone might conflate ``hard'' with ``unreached
on the stochastic axis.'' The deterministic axis, usually
treated as a degenerate corner case (``$k\!=\!1$ greedy'') or a
fallback, is doing more work here than typically credited: when
constructed via residual-stream interventions rather than ordinary
decoding, it exposes a slice that no sampling seed in
$\{42,\dots,47\}$ reaches. Grafting is not a decoding method; the point is that this slice is structurally identifiable, not that the unmodified model reaches it under ordinary inference. The fix is one extra deterministic forward pass per example plus a few cheap perturbations. We use activation grafting as a diagnostic tool to diversify the deterministic coverage and to locate this slice.

\textbf{Why this matters for mathematical reasoning.} 
Math reasoning is the setting where pass@$k$-style difficulty labels are most consequential, and where our effect is largest
in absolute terms. (i) \emph{Benchmark construction:} hardness
buckets derived from pass@$k\!=\!0$ are used to advertise
benchmark difficulty and to compare frontier models; a
structurally identifiable
$10.3$--$22.9\%$ of the ``hardest'' GSM8K / MATH bucket is in fact
reached by an edited forward pass of the same model under a cheap
deterministic diversification, so the bucket is partly an artifact
of the decoding regime, not of the items.
(ii) \emph{Difficulty calibration:} curricula and per-example
weights derived from sample-only pass-rates downweight or drop
items that are reachable off the stochastic axis under
residual-stream interventions; this slice is
non-trivial precisely on the hardest items, where calibration
matters most.
(iii) \emph{Synthetic curriculum and data curation:} rejection-sampling
pipelines (e.g., \citet{yuan2023rft,toshniwal2024openmathinstruct})
filter math problems by whether any sampled chain solves them; a
fraction of what those pipelines discard as ``unsolvable by the
generator'' is actually solvable by it at matched compute.
(iv) \emph{Verifier and reward-model training:} preference and
verifier datasets built from sampled-chain correctness inherit
the same blind spot; items with no positive sample contribute only negatives, even when a deterministic chain at matched compute would produce a positive. We do not propose to deploy activation grafting downstream; we use it to expose this slice
so that math-reasoning pipelines account for it.

\section{Related Work}
\textbf{Self-consistency and inference-time scaling.}
Self-consistency aggregates $k$ sampled chains by majority
vote~\citep{wang2022selfconsistency} and is the default protocol
for inference-time scaling on reasoning
benchmarks~\citep{wang2022selfconsistency,chen2023universal,aggarwal2023letscalmly}. Variants
adaptively allocate the budget~\citep{aggarwal2023letscalmly},
weight chains by self-evaluation~\citep{chen2023universal}, or
combine sampling with verifier-guided search~\citep{cobbe2021gsm8k,lightman2023letsverifystepbystep}.
Across this line of work, $k\!=\!1$ greedy is reported as a weak baseline that majority-voted sampling improves on. We do not dispute the aggregate ranking; we point out that the cell where greedy is correct and sampling-majority is not is non-trivial and
persistent, and is hidden by aggregate accuracy.

\textbf{Decoding diversity.}
Existing methods for decoding diversity operate in token space:
nucleus and top-$k$ sampling~\citep{holtzmancurious,fan2018hierarchical},
diverse beam
search~\citep{vijayakumar2016diverse,kool2019stochastic}, and
contrastive decoding~\citep{li2023contrastive}; see \citet{welleck2024meta} for a survey of inference-time decoding and meta-generation. All inject
diversity through the per-step output distribution. Our diagnostic is orthogonal: the pass@$k\!=\!0$ stratum persists at $k\!=\!6$ under
standard self-consistency hyperparameters, and is reached by
deterministic perturbations applied to the residual stream rather
than the output distribution.

\textbf{Hidden-state and representation interventions.}
A growing body of work edits or analyzes internal
activations: causal tracing for factual
recall~\citep{meng2022locating}, representation engineering and
steering~\citep{zou2023representation,turner2023activationaddition,rimsky-etal-2024-steering},
and cross-model communication via projected
hidden states~\citep{ramesh2025communicating}. We borrow this
primitive only as a probe, a minimal residual-stream perturbation
whose deterministic outputs we compare against sampling, not as a method that should be deployed at inference.

\textbf{Test-time compute and difficulty estimation.}
Recent work on test-time compute scaling reports favourable
exchange rates between extra inference passes and accuracy on
reasoning benchmarks~\citep{snell2024scaling,brown2024large,wu2024empirical}.
A common companion measure scores per-example difficulty by
pass@$k$ or sampling agreement~\citep{cobbe2021gsm8k,brown2024large},
and the same signal is used to filter hard subsets out of a larger dataset~\citep{zhou2025cats}.
Downstream RL-with-verifiable-rewards
pipelines~\citep{lambert2024tulu3,deepseekai2025r1} consume this
signal: prompts on which the model never samples a correct answer
contribute no reward gradient.
Our second observation directly affects this practice: a
non-trivial fraction of examples that fail under sampling are
solved by deterministic decoding, so sampling-only difficulty
estimates conflate ``hard'' with ``unreachable on the stochastic
axis''. Prior work on inference-time scaling has measured how coverage of correct answers grows with the sampling
budget~\citep{brown2024large}, and complementary theoretical
analyses~\citep{levi2025simple,levi2026learningshrinkshardtail}
characterise pass@$k$ as exhibiting a power-law decay in $k$ whose
exponent depends on the interaction between per-example difficulty
and training; \citet{levi2026learningshrinkshardtail} further shows
that learning ``shrinks the hard tail'' of the failure distribution
until an irreducible target-difficulty limit is reached. Our angle
is orthogonal: rather than modelling how the pass@$k$ failure rate
shrinks with $k$ under sampling, we ask what the residual
pass@$k\!=\!0$ stratum \emph{actually contains}, and isolate a
structurally identifiable slice of it that a matched-compute
deterministic regime reaches while the sampling axis does not.

\section{Setup}
\label{sec:setup}

We evaluate on the cross product of four open-weight instruction-tuned
models, \texttt{Qwen-2.5-3B}~\citep{qwen25},
\texttt{Llama-3.2-3B},
\texttt{Llama-3.1-8B}~\citep{llama32}, and
\texttt{Mistral-Nemo-12B}~\citep{mistralnemo}, and three reasoning
benchmarks, GSM8K~\citep{cobbe2021gsm8k},
MATH~\citep{hendrycks2021math} and MMLU-Pro~\citep{wang2024mmlupro};
$n=1000$ matched prompts per cell, and \texttt{max\_new\_tokens}$=2048$ to prevent truncation.

\begin{figure}[t]
\centering
\includegraphics[width=\linewidth]{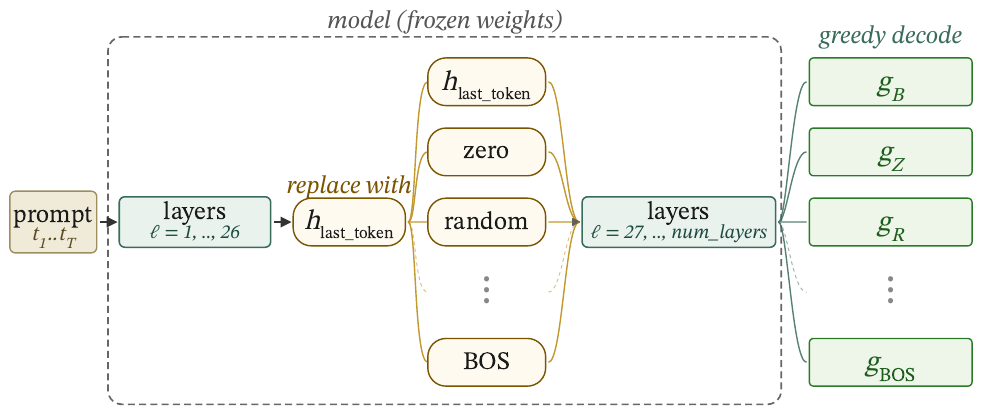}
\caption{\emph{Constructing the deterministic regime.}
The frozen model is run on a prompt up to layer~$26$, producing the
last-prompt-token residual-stream activation $h_{\text{last\_token}}$. It is then \emph{replaced} with one of $k$ cheap, parameter-free alternatives: $h_{\text{last\_token}}$ itself (the greedy baseline), the zero vector, a Gaussian
sample, \ldots, the BOS-token activation. The same model continues its forward pass through layers~$27$ to \texttt{num\_layers} under greedy decoding. The $k$ chains differ only in this one residual vector, yielding $k$ deterministic outputs $g_B, g_Z, g_R, \dots, g_{\text{BOS}}$.}
\label{fig:grafting}
\end{figure}

\paragraph{Decoding regimes.}
The \emph{sampling regime} draws six chains $S_{1\dots 6}$ at $T=0.7$,
$p_{\text{top}}=0.9$; we report sample-level
quantities (pass@$k$, single-sample mean) directly, and where a sample aggregate is needed we use majority vote with ties broken uniformly at
random over $50$ seeds. A temperature sweep at $T\!\in\!\{0.3,0.7\}$
appears in \S\ref{sec:robustness} (Table~\ref{tab:temperature}).
The \emph{deterministic regime} consists of greedy chains that differ only in a single last-prompt-token activation graft at $\ell\!=\!26$, fired during prefill only; decoding is greedy throughout. All graft vectors are fixed across the dataset (no per-example optimisation, parameter-free, no extra forward passes beyond the prefill hook). Figure~\ref{fig:grafting} provides an overview of grafting, while table~\ref{tab:chains} lists the eight chains used in the paper.

\paragraph{Matched compute.}
By \emph{matched compute} we mean $k$ deterministic chains against $k$
sampled chains: the same number of prefill+decode forward passes per
example. The graft is a single \texttt{register\_forward\_hook} that
overwrites one residual-stream vector during prefill (no extra
forward pass, no gradient, negligible wall-clock overhead), so we
operationalise ``compute'' as the FLOP count of those $k$ passes. The
headline claim of Sec.~\ref{sec:diversify} is at $k\!=\!6$.

\begin{table}[h]
\centering\footnotesize
\caption{Deterministic chains used in the paper. Short names are used
interchangeably with the $g_\cdot$ symbols from \S\ref{sec:diversify}
onward.}
\label{tab:chains}
\setlength{\tabcolsep}{4pt}
\begin{tabular}{l|l|l}
\toprule
Symbol / name & Description & Set \\
\midrule
$g_B$           & greedy, no graft               & 3 \\
$g_Z$ / \textsc{zero}  & zero vector            & 3 \\
$g_R$ / \textsc{rand}  & norm-matched Gaussian  & 3 \\
\textsc{runit}  & random unit direction          & 6 \\
\textsc{shuf}   & hidden-dim permutation of $g_B$'s activation & 6 \\
\textsc{bos}    & BOS-token activation           & 6 \\
\textsc{avg}    & mean over prompt-token activations & 8 \\
\textsc{prev}   & same position from $\ell\!-\!1$ & 8 \\
\bottomrule
\end{tabular}
\end{table}

Chains were chosen to be mechanistically distinct in residual-stream geometry (the origin, a high-norm random direction, a token-position
substitution, a hidden-dim permutation, a previous-layer state). We
quantify the resulting fix-set diversity in \S\ref{sec:diversify}
(cross-kind Jaccard $\leq\!0.47$ in every cell vs.\ a same-axis
ceiling reaching $0.76$ between two random-direction probes;
App.~\ref{app:extragraft}). Layer and position choices shift the oracle
ceiling by $\leq\!4$pp and reorder leading methods
(Apps.~\ref{app:layers}, \ref{app:position}, \ref{app:extragraft}); the existence of the recovered slice is insensitive to these choices, the specific chain that recovers a given example is not.

\section{Greedy is Competitive at $k\!=\!1$}
\label{sec:greedy-vs-mean}

At $k\!=\!1$, the sampling regime is one stochastic draw $S_i$, the
deterministic regime is $g_B$. Running six independent sampling seeds
($42$--$47$), greedy is competitive with or better than the mean
single-sample accuracy (Table~\ref{tab:k1}), with the gap reaching
up to $+2.8\%$ on \texttt{Llama-3.2-3B} / MATH. On six of twelve cells it
exceeds the seed-to-seed range and beats the best seed. Greedy follows the local MAP trajectory while a single
sample is a high-variance estimator of it; on harder problems the
per-step distribution is flatter and MAP becomes a better single-pass
bet than any draw.

\begin{table}[h]
\centering\scriptsize
\caption{Single-pass head-to-head over six sampling seeds.}
\label{tab:k1}
\begin{tabular}{l@{\hskip 8pt}|cc@{\hskip 8pt}c}
\toprule
Setup & $1S$ Mean [Min,Max] & $g_B$ & $\Delta$ \\
\midrule
Qwen-2.5-3B/GSM8K       & $0.842\,[0.838, 0.847]$ & 0.843 & $+0.001$ \\
Qwen-2.5-3B/MATH        & $0.544\,[0.539, 0.547]$ & 0.557 & $+0.013$ \\
Qwen-2.5-3B/MMLU-Pro    & $0.403\,[0.398, 0.411]$ & 0.401 & $-0.002$ \\
\midrule
Llama-3.2-3B/GSM8K      & $0.748\,[0.736, 0.756]$ & 0.770 & $+0.022$ \\
Llama-3.2-3B/MATH       & $0.461\,[0.448, 0.469]$ & 0.489 & $+0.028$ \\
Llama-3.2-3B/MMLU-Pro   & $0.345\,[0.336, 0.361]$ & 0.365 & $+0.020$ \\
\midrule
Llama-3.1-8B/GSM8K      & $0.837\,[0.826, 0.848]$ & 0.847 & $+0.010$ \\
Llama-3.1-8B/MATH       & $0.485\,[0.477, 0.492]$ & 0.485 & $+0.000$ \\
Llama-3.1-8B/MMLU-Pro   & $0.483\,[0.464, 0.502]$ & 0.496 & $+0.013$ \\
\midrule
Mistral-Nemo-12B/GSM8K  & $0.847\,[0.842, 0.851]$ & 0.857 & $+0.010$ \\
Mistral-Nemo-12B/MATH   & $0.321\,[0.307, 0.340]$ & 0.340 & $+0.019$ \\
Mistral-Nemo-12B/MMLU-Pro & $0.431\,[0.418, 0.437]$ & 0.442 & $+0.011$ \\
\bottomrule
\end{tabular}
\end{table}

The deterministic regime's clearest per-pass advantage is at $k\!=\!1$;
at higher $k$, its contribution is better understood in terms of coverage
(Sec.~\ref{sec:diversify}).

\section{The Blind Spot is Structural, Not Hardness}
\label{sec:diversify}

\paragraph{The stratum.}
The pass@$k{=}0$ stratum, examples no sampled chain in $k$
reaches gold, is the operative signal for label-required
pipelines (RL with verifiable rewards, data curation, hardness
annotation). At $k\!=\!6$ this stratum holds $5.1$--$43.5\%$ of
prompts ($51$ to $435$ examples per setup; see
Table~\ref{tab:Rk}). This is the slice
Sec.~\ref{sec:diversify} attacks at matched compute.

\begin{figure}[t]
\centering
\includegraphics[width=0.9\linewidth]{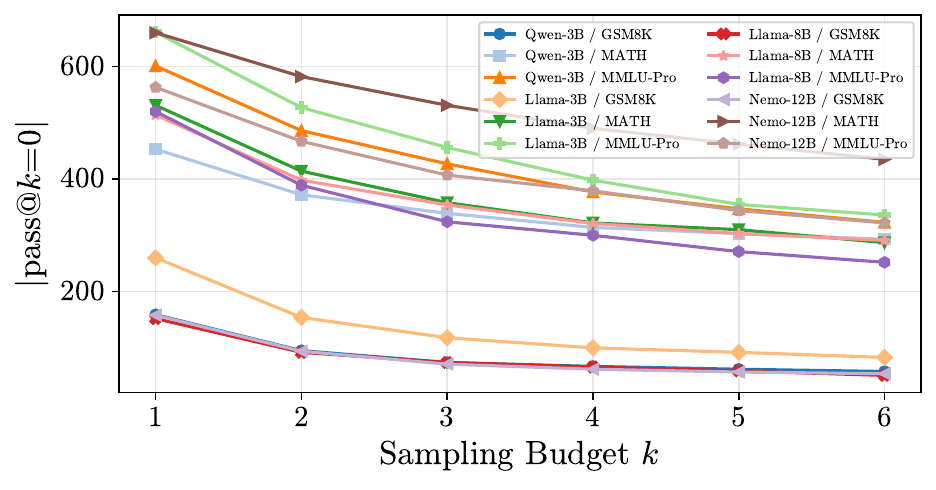}
\caption{Size of the pass@$k\!=\!0$ stratum as a function of
sampling budget $k$, across the twelve (model, benchmark) cells
($n\!=\!1000$ each). The curve flattens early: most of the shrink from $k\!=\!1$ to $k\!=\!6$ happens by $k\!=\!3$, and a residual
$5.1$--$43.5\%$ of prompts remains unreached at $k\!=\!6$. The
diminishing returns motivate spending budget on a different
axis.}
\label{fig:shrink}
\end{figure}

\paragraph{The stratum persists at higher $k$.}
The per-sample marginal-shrink rate of pass@$k\!=\!0$ decreases
monotonically across all twelve setups (Fig.~\ref{fig:shrink};
App.~\ref{app:robustness}). Within our sampling budget the shrinkage
has already plateaued: pass@$6\!=\!0$ still contains $5.1$--$43.5\%$
of prompts, and additional samples do not reliably close the gap.
The stratum we attack is therefore not an artifact of an undersized
sampling denominator.

Two readings of the residual
slice are possible: those examples are intrinsically hard, or
they are reachable but not on the stochastic axis.
Section~\ref{sec:diversify} adjudicates between them.

\paragraph{Anatomy.} The recovered pass@$6\!=\!0$ examples share a
recurring structure: the prompt admits two internally coherent
readings of a discrete choice, sampling drifts toward one reading
on all six seeds, and at least one deterministic chain commits to
the gold reading. Which chain rescues a given example varies by
example: no single graft dominates, but the union over a handful
of mechanistically distinct chains catches a non-trivial fraction
of the stratum. As a representative case, MATH Counting\,\&\,Probability \texttt{0445} on \texttt{Qwen-2.5-3B} asks: \emph{``A mother
purchases 5 blue plates, 2 red plates, 2 green plates, and 1
orange plate. How many ways are there for her to arrange these
plates for dinner around her circular table if she doesn't want
the 2 green plates to be adjacent?''} (gold: $588$). Greedy and
the random graft both account for the identical-colour
multiplicities and the circular symmetry, computing
$10!/(5!\,2!\,2!\cdot 10) - 9!/(5!\,2!\cdot 9) = 756 - 168 = 588$.
The zero graft instead treats the plates as $9$ distinct objects
and returns $9! - 8! \cdot 2 = 282{,}240$, ignoring both the
identical-colour multiplicities and the circular division. All six sampled seeds in $\{42,\dots,47\}$ return non-gold
answers, so the item is in pass@$6\!=\!0$, yet two of the three
deterministic chains recover it. Two further worked examples,
spanning the other two rescuing pairs from $\{g_B,g_Z,g_R\}$, are
in App.~\ref{app:cases}.

\paragraph{Adjudicating the two readings.}
To adjudicate between the two readings of the residual sampling-failure slice, we work directly on the pass@6$=$0 stratum: examples for which
no seed in $\{42,\dots,47\}$ reaches gold. On this slice, no sampled
chain in six tries was correct: this is the strictest
sampling-failure test. The slice contains $51$ to $435$ examples per
setup (Table~\ref{tab:Rk}). We then ask: of these, how many does a
deterministic regime at matched compute reach? Those would not actually be hard, but anyway flagged as hard by a sampling-based filter. 

We define $R_k$ as the number of pass@6$=$0 examples on which at
least one of $k$ deterministic chains is correct. $R_1$ is greedy
alone. $R_3$ adds zero and norm-matched random grafts ($g_Z, g_R$).
$R_6$ extends to greedy plus five grafts (zero, random, random unit
direction, hidden-dim permutation, BOS-token replacement),
matching the $k\!=\!6$ sampling budget. $R_8$ adds mean-activation and
previous-layer grafts. All grafts are at $\ell\!=\!26$, last token position.
The headline matched-compute claim is $R_6$ vs.\ pass@6$=$0 (six
deterministic chains against six sampled chains, on the stratum
where no sample reaches gold); $R_1$ and $R_3$ are reported at
sub-matched budget and so strengthen, not weaken, the comparison
(one or three deterministic chains reaching examples that six
samples miss).

\begin{table*}[h]
\centering \footnotesize
\caption{Deterministic recovery on the pass@6$=$0 slice. $R_k$ counts
examples where no sampling seed in $\{42,\dots,47\}$ reaches gold but
at least one of $k$ deterministic chains does. Percentages are of the
pass@6$=$0 slice (column 2); $n=1000$ for all cells.}
\label{tab:Rk}
\begin{tabular}{l|c|cccc}
\toprule
Setup & p@6$=$0 & $R_1$ & $R_3$ & $R_6$ & $R_8$ \\
\midrule
\texttt{Qwen-2.5-3B} / GSM8K       &  58 &  0 (\phantom{0}0.0\%) &  5 (\phantom{0}8.6\%) &  8 (13.8\%) & 10 (17.2\%) \\
\texttt{Qwen-2.5-3B} / MATH        & 293 & 10 (\phantom{0}3.4\%) & 23 (\phantom{0}7.8\%) & 39 (13.3\%) & 48 (16.4\%) \\
\texttt{Qwen-2.5-3B} / MMLU-Pro    & 323 & 14 (\phantom{0}4.3\%) & 43 (13.3\%) & 81 (25.1\%) & 100 (31.0\%) \\
\midrule
\texttt{Llama-3.2-3B} / GSM8K      &  83 &  2 (\phantom{0}2.4\%) & 12 (14.5\%) & 19 (22.9\%) & 21 (25.3\%) \\
\texttt{Llama-3.2-3B} / MATH       & 287 & 11 (\phantom{0}3.8\%) & 24 (\phantom{0}8.4\%) & 42 (14.6\%) & 50 (17.4\%) \\
\texttt{Llama-3.2-3B} / MMLU-Pro   & 336 & 30 (\phantom{0}8.9\%) & 66 (19.6\%) & 96 (28.6\%) & 114 (33.9\%) \\
\midrule
\texttt{Llama-3.1-8B} / GSM8K      &  51 &  3 (\phantom{0}5.9\%) &  7 (13.7\%) & 10 (19.6\%) & 11 (21.6\%) \\
\texttt{Llama-3.1-8B} / MATH       & 291 &  9 (\phantom{0}3.1\%) & 22 (\phantom{0}7.6\%) & 30 (10.3\%) & 36 (12.4\%) \\
\texttt{Llama-3.1-8B} / MMLU-Pro   & 252 & 19 (\phantom{0}7.5\%) & 46 (18.3\%) & 72 (28.6\%) & 80 (31.7\%) \\
\midrule
\texttt{Mistral-Nemo-12B} / GSM8K  &  54 &  1 (\phantom{0}1.9\%) &  8 (14.8\%) & 11 (20.4\%) & 13 (24.1\%) \\
\texttt{Mistral-Nemo-12B} / MATH   & 435 & 16 (\phantom{0}3.7\%) & 39 (\phantom{0}9.0\%) & 64 (14.7\%) & 75 (17.2\%) \\
\texttt{Mistral-Nemo-12B} / MMLU-Pro & 322 & 11 (\phantom{0}3.4\%) & 55 (17.1\%) & 93 (28.9\%) & 109 (33.9\%) \\
\bottomrule
\end{tabular}
\end{table*}

Three observations follow. (a)~At matched compute, $R_6$ recovers
$10$--$29\%$ of examples that six sampling seeds collectively miss.
The pass@6$=$0 stratum is therefore not intrinsically hard: a
non-trivial fraction is reachable by deterministic decoding given
enough representatives. (b)~$R_1$ is still relatively small ($0$--$9\%$),
so a single deterministic chain is closer to
``another sample'' than to a separate regime; the regime emerges once multiple mutually disjoint perturbations are admitted. The
recovery scales near-linearly with the deterministic budget
($R_6$ is several times $R_1$ on every setup where $R_1>0$). (c)~The recovery is not driven
by one privileged vector. Among the seven grafts surveyed, no
single graft kind is the single largest individual contributor to
$R_6$ on more than half the cells, and the leader rotates across
cells (\textsc{shuffled} on the math-style cells we inspected,
\textsc{random} on \texttt{Mistral-Nemo-12B} / MMLU-Pro, multi-way ties on the
smallest strata). Across all twelve cells, pairwise cross-kind
fix-set Jaccard never exceeds $0.47$ (range $0.06$--$0.47$;
App.~\ref{app:extragraft}, Table~\ref{tab:fixset_jaccard_all}),
while the single same-axis pair (random vs.\ random-unit, two
norm-matched Gaussians) reaches $0.76$ on \texttt{Qwen-2.5-3B} / GSM8K as an
illustrative ceiling for what resampling along a single random
axis can contribute (see App.~\ref{app:extragraft} for the
per-cell breakdown).
Different graft kinds therefore probe different subsets of the
stratum in every cell we test, which is consistent with the
recovery scaling with the deterministic budget.

These three readings together rule out both the ``intrinsically
hard'' explanation and the ``one privileged vector'' explanation.
Crucially, the recovery on pass@6$=$0 \emph{scales with the
deterministic budget}, admitting more mutually disjoint
deterministic perturbations reaches more examples no sampling seed
reached, and the dimensions on which it scales are mechanistically
distinct from each other and from sampling's stochastic axis. What
looked intrinsically hard under sampling was, for a non-trivial
fraction, just unreached on the stochastic axis: the model was
close, the missing ingredient was a different \emph{kind} of
diversity.

$66$--$88\%$ of pass@6$=$0 remains unadjudicated by an 8-chain
deterministic regime: those examples may be intrinsically hard,
or reachable on some other diversity axis we do not probe. Within
the 3-chain set, each of $\{g_B, g_Z, g_R\}$ contributes some
examples uniquely (Table~\ref{tab:marginal}): no single chain
dominates across the twelve cells, and the deterministic budget
pays back per chain. Per-setup unique contributions sit in the
$1.3$--$8.9\%$ band of $n=1000$, with $g_B$ leading in nine of
twelve cells but never to the point of making either graft
redundant, removing any one chain drops the 3-chain oracle by at
least $13$ examples on every setup, and the leading and trailing
contributor differ by a factor of $\leq 2.2$.

\begin{table}[h]
\centering\scriptsize
\caption{Per-source unique contribution to the 3-chain deterministic
oracle $\det^\oplus = g_B \cup g_Z \cup g_R$ across all twelve cells.
``Unique to $x$'' counts examples whose gold answer is reachable
through $x$ but not the other two ($n=1000$); equivalently, the drop
in $\det^\oplus$ if $x$ is removed. No single chain dominates: $g_B$
leads on nine cells, $g_R$ on two (both \texttt{Llama-3.1-8B}), and $g_B$ is
tied with $g_Z$ on \texttt{Llama-3.2-3B} / GSM8K.}
\label{tab:marginal}
\setlength{\tabcolsep}{2.5pt}
\begin{tabular}{l|ccc}
\toprule
Setup & unique to $g_B$ & unique to $g_Z$ & unique to $g_R$ \\
\midrule
\texttt{Qwen-2.5-3B} / GSM8K       & 24 (2.4\%) & 21 (2.1\%) & 23 (2.3\%) \\
\texttt{Qwen-2.5-3B} / MATH        & 33 (3.3\%) & 30 (3.0\%) & 21 (2.1\%) \\
\texttt{Qwen-2.5-3B} / MMLU-Pro    & 68 (6.8\%) & 57 (5.7\%) & 58 (5.8\%) \\
\midrule
\texttt{Llama-3.2-3B} / GSM8K      & 37 (3.7\%) & 37 (3.7\%) & 31 (3.1\%) \\
\texttt{Llama-3.2-3B} / MATH       & 79 (7.9\%) & 37 (3.7\%) & 53 (5.3\%) \\
\texttt{Llama-3.2-3B} / MMLU-Pro   & 89 (8.9\%) & 61 (6.1\%) & 58 (5.8\%) \\
\midrule
\texttt{Llama-3.1-8B} / GSM8K      & 14 (1.4\%) & 13 (1.3\%) & 23 (2.3\%) \\
\texttt{Llama-3.1-8B} / MATH       & 55 (5.5\%) & 39 (3.9\%) & 53 (5.3\%) \\
\texttt{Llama-3.1-8B} / MMLU-Pro   & 49 (4.9\%) & 54 (5.4\%) & 57 (5.7\%) \\
\midrule
\texttt{Mistral-Nemo-12B} / GSM8K  & 35 (3.5\%) & 23 (2.3\%) & 21 (2.1\%) \\
\texttt{Mistral-Nemo-12B} / MATH   & 71 (7.1\%) & 44 (4.4\%) & 45 (4.5\%) \\
\texttt{Mistral-Nemo-12B} / MMLU-Pro & 77 (7.7\%) & 52 (5.2\%) & 71 (7.1\%) \\
\bottomrule
\end{tabular}
\end{table}

\section{Mechanism: Residual-Stream Perturbation, Not Attention Rerouting}
\label{sec:mechanism}

Counts and case studies establish that grafts reach examples
sampling does not. They do not, on their own, explain why the
deterministic chains in our table behave as multiple distinct axes
rather than a single noisy one. This section gives the
mechanistic answer.

We instrumented \texttt{Qwen-2.5-3B} with eager attention on $100$
GSM8K examples and ran each twice: once unperturbed, once with a
last-token graft at $\ell\!=\!26$ (random and zero, separately).
For every transformer block we recorded (i) the hidden state at
the graft position before and after grafting, yielding cosine
similarity and $L_2$ distance per layer, and (ii) the full
attention-weight matrices, allowing direct comparison between
baseline and grafted runs at identical token positions. For the
random graft, $4/100$ examples exhibited a runaway-divergence mode
in which $L_2$ grows to $10^4$--$10^5$ in deeper layers; we exclude
these from the per-layer summary to avoid heavy-tail contamination
and report the trajectory on the remaining $96$. Zero shows no
such outliers.

\begin{figure}[t]
\centering
\includegraphics[width=0.94\columnwidth]{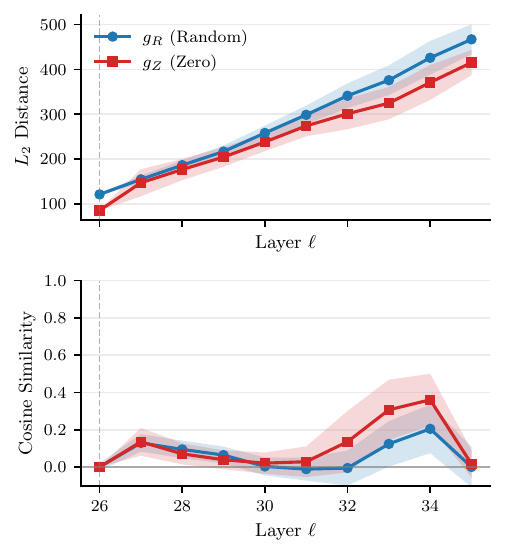}
\caption{Hidden-state divergence at the graft position over layers
$\ell \geq 26$ (\texttt{Qwen-2.5-3B}, GSM8K, last token, $\ell\!=\!26$
graft; random excludes 4 runaway-divergence outliers, $n=96$).
Cosine similarity to the unperturbed run stays low and $L_2$ distance
grows monotonically through the remaining nine blocks: the
perturbation is carried forward, not absorbed. Baseline norm at the
graft position is $86.2 \pm 0.9$ at $\ell\!=\!26$ and $309.2 \pm 14.5$
at $\ell\!=\!35$.}
\label{fig:mech_traj}
\end{figure}

Hidden-state divergence is shown in Figure~\ref{fig:mech_traj}
(per-layer numbers in App.~\ref{app:mech_table},
Table~\ref{tab:mech_traj}). At $\ell = 26$, the random graft makes
the hidden state essentially orthogonal to the baseline
(cos $= 0.001$) at $L_2$ distance $121.8$; the zero graft sets it
to the origin ($L_2 = 86.2$, the baseline norm itself). Crucially
the divergence does \emph{not} decay through the remaining nine
layers: $L_2$ grows monotonically to $466.7$ (random) and $415.1$
(zero) by the final layer, while cosine similarity to baseline
remains below $0.37$. In this model, the forward computation
does not project the perturbation back onto the baseline
trajectory; it carries it forward.

By contrast, the graft's \emph{direct} effect on attention is
negligible (attention over the divergent generation itself is not
what we measure here; see footnote). We report two
quantities: (a) the mean absolute difference in
\texttt{prefill\_attn\_received} (per-token attention received
during the prompt forward pass, averaged over heads and query
positions) between baseline and grafted runs, and (b) the mean
absolute difference in the full attention-weight matrix under
teacher forcing on \emph{identical} token sequences (prompt +
baseline output for both runs, so any difference is attributable
to the graft and not to divergent generation).\footnote{Teacher
forcing the grafted run on the baseline output is necessary to
isolate the graft's direct effect on attention from indirect
effects mediated by divergent generation: the grafted chain
typically produces a different output, and attention over that
different output would conflate the two. Comparing on the
\emph{same} token sequence asks the cleaner question: does
the graft, at layer $\ell\!=\!26$, change which
tokens subsequent attention heads attend to?} On the $100$
examples, (a)$\,= (1.1 \pm 0.3)\!\times\!10^{-5}$ and
(b)$\,= (3.5 \pm 0.8)\!\times\!10^{-5}$ for the random graft; the
zero graft is within rounding ($(1.2 \pm 0.4)\!\times\!10^{-5}$ and
$(3.5 \pm 0.9)\!\times\!10^{-5}$). This is four to five orders of
magnitude smaller than the residual-stream perturbation the same
intervention induces.

\paragraph{From mechanism to behaviour.}
The two measurements above pin down how the graft acts on the forward
pass: it does not reroute attention but injects a content vector that
the remaining nine blocks carry forward without correction
(Figure~\ref{fig:mech_traj}, $L_2$ grows monotonically; attention
deltas are $10^{-5}$). Mechanistically, each graft vector is therefore
a different initial condition for the same nine-block computation
producing the first generated token, and because the forward
pass does not contract these perturbations back onto the baseline
trajectory, distinct graft kinds produce distinct downstream
trajectories that the autoregressive decode then locks in. This is the link to the
behavioural table: different residual-stream perturbations propagate
into different generated answers, and the eight chains in
Table~\ref{tab:Rk} are eight non-collinear initial conditions, not
eight redraws from one.

The deterministic axis is therefore multi-dimensional in a
behaviourally measurable sense: cross-kind fix-set Jaccard
$\leq\!0.47$ in every one of the twelve cells
(App.~\ref{app:extragraft}) shows the chains act on largely
disjoint subsets of the input distribution. That is why $R_k$
scales with the deterministic budget rather than saturating after
one graft: more graft kinds are more independent initial
conditions, each reaching its own slice of pass@6$=$0.

\section{Practical Utility}
\label{sec:deployment}

The headline finding of \S\ref{sec:diversify} is qualitative: a
six-chain deterministic regime reaches $10$--$29\%$ of pass@$6\!=\!0$.
This section turns it into two \emph{deployable} recipes that need
neither gold labels nor per-cell tuning. First, a matched-cost
substitution (\S\ref{sec:matched_swap}): a fixed-policy rule that
spends one of $B$ forward passes on the mean-prompt-activation graft
$\textsc{avg}$ rather than an extra sample covers more of the dataset
than $B$ samples on the majority of (cell, budget) combinations, with
$\textsc{avg}$ chosen once and never tuned. Second, a label-free
curation flag (\S\ref{sec:label_free}): chain disagreement among
$\{g_B, S_0, S_1\}$ ranks recoverable items above the base rate
without gold labels, with $3$--$5\times$ lift on free-form math at
three forward passes per item. Both recipes are deployable directly
on top of an existing pass@$k$ curation pipeline, at the cost of a
small constant overhead per item.

\subsection{Matched-cost coverage}
\label{sec:matched_swap}

A sharper test of the \S\ref{sec:diversify} finding is direct
substitution at matched compute: at total budget $B$ forward passes
per item, does $(B\!-\!1)$ samples plus \emph{one} deterministic
chain cover more of the dataset than $B$ samples? We propose the
mean-prompt-activation graft $\textsc{avg}$ as a fixed, no-tuning
rule: at every budget and every cell, spend one of the $B$
forwards on $\textsc{avg}$ rather than an extra sample. Across the
$96$ (cell, $B$) slots with $B \in \{2,\dots,9\}$, the
fixed-$\textsc{avg}$ policy reduces residual pass@$B\!=\!0$ count
by $+1.66$ items per slot on average, and is net-positive on
$8$ of $12$ cells: all four MMLU-Pro cells (mean $\Delta$
$+4.4$ to $+7.0$ per slot), \texttt{Qwen-2.5-3B} / MATH ($+7.9$),
\texttt{Mistral-Nemo-12B} / MATH ($+4.5$), \texttt{Llama-3.1-8B} / GSM8K ($+2.0$),
and \texttt{Mistral-Nemo-12B} / GSM8K ($+0.4$).
As an upper bound, we also report the oracle in which the best
chain at each $(\text{cell},B)$ is picked from the eight in
Table~\ref{tab:chains}; this is an analysis, not a deployable
rule. The oracle wins on $80$ of $96$ slots and is particularly
large on MMLU-Pro and Mistral-Nemo cells (e.g.\
\texttt{Mistral-Nemo-12B} / MMLU-Pro: $379\!\to\!357$ at $B\!=\!4$ and
$277\!\to\!269$ at $B\!=\!9$; \texttt{Llama-3.2-3B} / MMLU-Pro:
$336\!\to\!317$ at $B\!=\!6$; \texttt{Qwen-2.5-3B} / MATH: $-10$ at
$B\!=\!6$). The unconstrained best mix even goes det-heavy on
three cells at $B\!=\!9$ ($s\!=\!1,d\!=\!8$ on
\texttt{Qwen-2.5-3B} / MATH; $s\!=\!2,d\!=\!7$ on \texttt{Mistral-Nemo-12B} /
MMLU-Pro; $s\!=\!4,d\!=\!5$ on \texttt{Llama-3.2-3B} / GSM8K). The gap
between fixed-$\textsc{avg}$ and the oracle is modest at the slot
level (median best-chain wins $\approx\!1$--$2$ items more than
$\textsc{avg}$ when they differ), so $\textsc{avg}$ reaches
near-oracle coverage with no per-cell tuning.
Per-cell, per-$B$ oracle tables are in
App.~\ref{app:matched_swap}. The headline is operational: when
budgeting forward passes for coverage on a curation task,
spending one of them on the $\textsc{avg}$ graft rather than an
extra sample is the better choice on the majority of
(cell, budget) combinations. The choice of a \emph{single} graft
slot is a proof-of-concept simplification: the same matched-cost
substitution extends to $(B\!-\!d)$ samples plus $d$ deterministic
chains for any $d\!\leq\!B$, and a curation pipeline with a fixed
audit budget per item can tune $d$ to its own
sample/deterministic exchange rate.

\subsection{Toward label-free identification}
\label{sec:label_free}

A natural follow-up is whether the recoverable slice can be
flagged \emph{without} gold labels, purely from chain agreement.
As a deployable probe we use only three forward passes per item, the greedy chain $g_B$ and two sampling seeds $S_0, S_1$, and score each item by how many of the three pairs disagree.
We then rank items by this score and ask whether the top-$K$\%
most-suspect items contain a higher density of recoverable items
(pass@$6\!=\!0$ \emph{and} some deterministic chain correct)
than random selection would.
The base rate of recoverable items is $0.8$--$9.6\%$ across the
twelve cells. On the eight free-form math cells (GSM8K, MATH) the
probe concentrates recoverable items at the top of the ranking, with
lift $3$--$5\times$ at $K\!=\!2$--$5\%$ on those cells and precision up to
$20\%$ at $K\!=\!1\%$ on the strongest cell (\texttt{Llama-3.1-8B} / MATH);
lift declines with $K$, as expected for a ranker. The signal lowers with MMLU-Pro cells, where the lift lies within
$1.0$--$1.5\times$ (see Limitations). Aggregate ``$3$--$5\times$'' is
therefore a majority claim, not universal.
Per-setup numbers are in
App.~\ref{app:label_free}, Table~\ref{tab:label_free};
Fig.~\ref{fig:diff_flag} shows precision-recall curves.
This reframes the deployment from an accuracy intervention
into a curation tool: given a fixed budget to re-examine $K\%$
of items with deterministic perturbations, the probe identifies
the $K\%$ most likely to be \emph{mislabelled} as intrinsically
hard, at three forward passes per item and no gold labels.
A calibrated precision guarantee (rather than rank-only lift), however, 
would require a small labelled dev set per pipeline, which we
leave to future work.

\section{Discussion}
\label{sec:discussion}

\paragraph{Robustness.}
\label{sec:robustness}
Mechanism is residual-stream perturbation, not attention rerouting
(§\ref{sec:mechanism}); the diagnostic holds at lower sampling
temperature (Table~\ref{tab:temperature}). On \texttt{Qwen-2.5-3B} / GSM8K with
$T\!=\!0.3$ across four seeds, the pass@$4\!=\!0$ stratum
\emph{grows} ($6.7\%\!\to\!8.2\%$ of $n$) and the deterministic
recovery on that stratum grows with it: $R_6$ rises from $20.9\%$
at $T\!=\!0.7$ to $31.7\%$ at $T\!=\!0.3$. Lowering $T$ peaks the
per-step distribution and reduces stochastic exploration, so the
sampling-only blind spot widens while the deterministic regime
continues to reach into it. The blind spot is therefore not a
high-temperature artifact; if anything, the diagnostic is more
visible at lower $T$.

\begin{table}[h]
\centering\small
\caption{Pass@$k\!=\!0$ and deterministic recovery on
\texttt{Qwen-2.5-3B} / GSM8K at two sampling temperatures
($n\!=\!1000$, four matched seeds each). $R_k$ counts examples
in the pass@$4\!=\!0$ stratum on which at least one of $k$
deterministic chains is correct; percentages are of the stratum.
Deterministic chains are temperature-independent; the only thing
that changes between columns is the sampling stratum.}
\label{tab:temperature}
\begin{tabular}{l | cc}
\toprule
& $T=0.3$ & $T=0.7$ \\
\midrule
pass@$4{=}0$ stratum size  & $82$ ($8.2\%$) & $67$ ($6.7\%$) \\
$R_1$ on stratum   & $3$ ($3.7\%$)  & $1$ ($1.5\%$) \\
$R_3$ on stratum   & $15$ ($18.3\%$) & $9$ ($13.4\%$) \\
$R_6$ on stratum   & $26$ ($31.7\%$) & $14$ ($20.9\%$) \\
\bottomrule
\end{tabular}
\end{table}

\paragraph{Intervention vs decoding.}
Activation grafting is an intervention on internal representations,
not a decoding method: it overwrites the activation the model actually
computed at layer~$26$ with a fixed vector and reads off predictions
from the edited forward pass. The two regimes we compare are therefore
not two ways of decoding the same system; one is the model running,
the other is an internally edited model at matched compute. Our claim
is correspondingly narrow: of the examples that pass@$k\!=\!0$
classifies as too hard, a non-trivial fraction is reached by an
edited forward pass at matched compute, so the slice is structurally
identifiable in the residual stream. We do not claim those items are
``truly easy'' or that the unmodified model reaches them under
ordinary inference; only that they are not unreachable in the broader
sense, and that their reachability is decoding-regime dependent rather
than an intrinsic property of the items. Relatedly, there is no
difficulty oracle: any operational notion of difficulty (pass@$k$,
sampling agreement, intervention-based reachability) is a choice, not
a measurement. The reverse asymmetry is real and we do not quantify
it: some items the deterministic regime misses are reachable by
sampling. We do not probe this direction because it is not what
downstream pipelines act on: sampling is the de facto default for
difficulty estimation, and pass@$k$-style proxies,
RL-with-verifiable-rewards filters, and curation pipelines are often
built on it. The deterministic regime, in current practice, is
collapsed to greedy alone --- a single chain, on which a pass@$k$
hardness signal is undefined --- so no operational pipeline currently
labels items hard based on deterministic misses. What needs auditing
is therefore the direction we report: items the sampling-based proxy
currently flags as hard, on a stratum that downstream pipelines treat
as such.

\paragraph{Limitations.}
Our study covers four open-weight models (3B, 8B and 12B parameters)
on three reasoning benchmarks (two free-form math, GSM8K and MATH,
plus one multiple-choice, MMLU-Pro), with seven graft vectors at a
single layer and position and a default sampling temperature of
$T\!=\!0.7$ (plus one $T\!=\!0.3$ check). Even with eight deterministic
chains, $66$--$88\%$ of the pass@$6\!=\!0$ stratum remains unrecovered;
those examples may be genuinely hard, or reachable along a diversity
axis we do not test. We report conditional recovery as point estimates
across the twelve cells, without bootstrap confidence intervals or
repeated-seed-set variance; the qualitative direction holds on every
cell, but per-cell rates should be read with this uncertainty in mind. We also do not study whether the recovered slice
can be identified \emph{without} gold labels at deployable precision;
we treat that as a separate question. The cheap cross-chain
disagreement probe in App.~\ref{app:label_free} works well on
free-form numeric and symbolic answers (lift@$5\%$ up to
$8.0\times$ on GSM8K / MATH cells) but degrades on multiple-choice
benchmarks (lift@$5\%$ collapses to $1.0$--$1.5\times$ across the four
MMLU-Pro cells): with a bounded answer vocabulary (A--J), two
sampled chains and greedy frequently agree by chance even on
recoverable items, so chain disagreement loses its signal. A
deployable label-free flag would need a different signal on
MCQ-style outputs. Finally, on the three smallest pass@$6\!=\!0$
strata (\texttt{Llama-3.1-8B} / GSM8K, $51$ examples, \texttt{Mistral-Nemo-12B} / GSM8K,
$54$, and \texttt{Qwen-2.5-3B} / GSM8K, $58$), absolute recovery counts are
small ($R_6\!\in\!\{10, 11, 8\}$); the recovery \emph{rates} line up
with the other cells, but the noise floor is higher.

\section{Conclusion}
Pass@$k$ is the canonical sampling-based primitive for per-example
difficulty, driving RL data curation, hardness annotation, and
curricula across reasoning-LM evaluation. On the twelve
(model, benchmark) cells we test (four open-weight models spanning
3B, 8B and 12B parameters across three reasoning benchmarks), we
show it has a persistent blind spot on its hardest stratum at the
$k\!=\!6$ budget: of examples no sampling seed in
$\{42,\dots,47\}$ solves, $10$--$29\%$ are reached by a six-chain
deterministic regime at matched compute, while greedy alone reaches
$0$--$9\%$. The recovery scales with the deterministic budget
across five graft kinds whose mechanistic distinctness we verify
across all twelve cells (cross-kind fix-set Jaccard $\leq\!0.47$
in every cell; App.~\ref{app:extragraft}). The decoder is
unchanged: stochastic decoding and sample majority remain the right default at inference,
and grafting is purely a diagnostic tool for locating the slice.

The implication is that sampling-derived difficulty annotations
conflate ``hard'' with ``unreached on the stochastic axis'' on a
non-trivial fraction of their hardest stratum, which is what
RL filters, curricula, and pass@$k$-based labels build on.
For AI-for-math pipelines specifically, benchmark hardness
buckets, pass@$k$-filtered synthetic curricula, and verifier
datasets, this means an identifiable fraction of items currently
labeled as too hard for the generator are, at matched compute,
not. The slice is structurally locatable and can be audited at
a few extra deterministic forward passes per item.

\section*{Acknowledgements}
This work is supported by the MUR FIS2 grant n. FIS-2023-00942 "NEXUS" (cup B53C25001030001), and partly by Sapienza University of Rome via the Seed of ERC grant "MINT.AI" (cup B83C25001040001).

\section*{Impact Statement}
This work is a diagnostic study of how pass@$k$-based hardness labels
behave on math and science reasoning benchmarks; it does not propose a
new inference method or a more capable model, and the activation-grafting
intervention is used only as an analysis tool rather than as something
to deploy. The main downstream consequence we anticipate is more careful
construction of math-reasoning evaluation suites, curricula, and
verifier datasets, where treating pass@$k\!=\!0$ as intrinsic hardness
can silently and erroneously drop a structurally identifiable slice of samples the
model can in fact solve. Better labels of this kind should reduce
wasted compute in rejection-sampling pipelines and reduce miscalibrated
difficulty signals in reinforcement-learning-with-verifiable-rewards
training. We do not foresee specific dual-use risks: the diagnostic
does not enable new generation capabilities, does not require any
training, and is run entirely on existing open-weight models and
publicly released benchmarks.

\bibliographystyle{icml2026}
\bibliography{refs}

\newpage
\appendix
\onecolumn

\section{Per-Layer Hidden-State Divergence (Numerical)}
\label{app:mech_table}

Table~\ref{tab:mech_traj} reports the per-layer mean and standard
deviation of cosine similarity and $L_2$ distance to the unperturbed
run, summarised graphically in Figure~\ref{fig:mech_traj}.

\begin{table}[h]
\caption{Hidden-state divergence at the graft position, mean $\pm$
std over $100$ GSM8K examples (\texttt{Qwen-2.5-3B}, last token,
$\ell\!=\!26$; random excludes 4 runaway-divergence outliers,
$n=96$). Layers below $26$ are not shown (cos $=1$, $L_2=0$
identically). The baseline norm at the graft position is
$86.2 \pm 0.9$ at $\ell = 26$ and $309.2 \pm 14.5$ at $\ell = 35$.}
\label{tab:mech_traj}
\centering\small
\begin{tabular}{c cc cc}
\toprule
       & \multicolumn{2}{c}{$g_R$ (random)} & \multicolumn{2}{c}{$g_Z$ (zero)} \\
$\ell$ & cos & $L_2$ & cos & $L_2$ \\
\midrule
26 & $0.001 \pm 0.021$ & $121.8 \pm 2.0$  & $0.000$ & $86.2 \pm 0.9$ \\
27 & $0.130 \pm 0.047$ & $155.2 \pm 8.2$  & $0.136 \pm 0.074$ & $147.7 \pm 29.9$ \\
28 & $0.095 \pm 0.047$ & $186.8 \pm 10.8$ & $0.071 \pm 0.056$ & $177.1 \pm 23.8$ \\
30 & $0.003 \pm 0.048$ & $258.4 \pm 16.2$ & $0.021 \pm 0.057$ & $238.3 \pm 20.3$ \\
32 & $-0.005 \pm 0.094$ & $341.3 \pm 27.5$ & $0.136 \pm 0.165$ & $301.2 \pm 34.3$ \\
34 & $0.205 \pm 0.130$ & $425.7 \pm 37.3$ & $0.361 \pm 0.139$ & $370.9 \pm 38.9$ \\
35 & $0.001 \pm 0.108$ & $466.7 \pm 33.4$ & $0.017 \pm 0.073$ & $415.1 \pm 28.1$ \\
\bottomrule
\end{tabular}
\end{table}

\section{Layer-Selection Sweep}
\label{app:layers}

We computed the $\det^\oplus$ oracle ceiling on \texttt{Qwen-2.5-3B} / GSM8K
at four layers $\ell \in \{2, 13, 26, 32\}$, spanning early, middle, late,
and final-block positions in the 36-layer stack. The ceilings are
$0.902$, $0.903$, $0.908$, and $0.905$, with corresponding
$G_1 = \det^\oplus - g_B$ values of $+5.9\%$, $+6.0\%$, $+6.5\%$, and $+6.2\%$.
$G_1$ remains within a $0.6\%$ band across this $30$-layer range:
the diversification effect is broadly layer-insensitive, consistent with
a perturbation propagating forward through the residual stream rather
than acting at a privileged depth. We adopt $\ell = 26$ as the marginal
best across the 3B main-text cells (sweep done on \texttt{Qwen-2.5-3B});
the same layer is used for the 8B cells, and the choice is not load-bearing.

\section{Pass@$k\!=\!0$ Shrink-Rate Diagnostics}
\label{app:robustness}

A natural concern is that six sampling chains undersample the
sampling regime, and that the pass@$k\!=\!0$ stratum would collapse
with more samples. Going from $k\!=\!4$ to $k\!=\!6$ shrinks pass@$k\!=\!0$
substantially in absolute terms, but the stratum remains
$5.1$--$43.5\%$ of prompts at $k\!=\!6$, and the per-sample
marginal-shrink rate is monotonically decreasing across the twelve
cells (Fig.~\ref{fig:shrink}). Within our $k\!\le\!6$ budget the
shrinkage has already plateaued, so the $R_6$ recoveries reported in
Sec.~\ref{sec:diversify} ($10$--$29\%$ of pass@6$=$0) are not a
quirk of an undersized sampling denominator. Whether the
stratum would eventually vanish at much larger $k$ (e.g.\
$k\!=\!50,100$) is outside the budget of this study.

\section{Recovery Scaling with the Deterministic Budget}
\label{app:rk_payoff}

\begin{figure}[h]
\centering
\includegraphics[width=\linewidth]{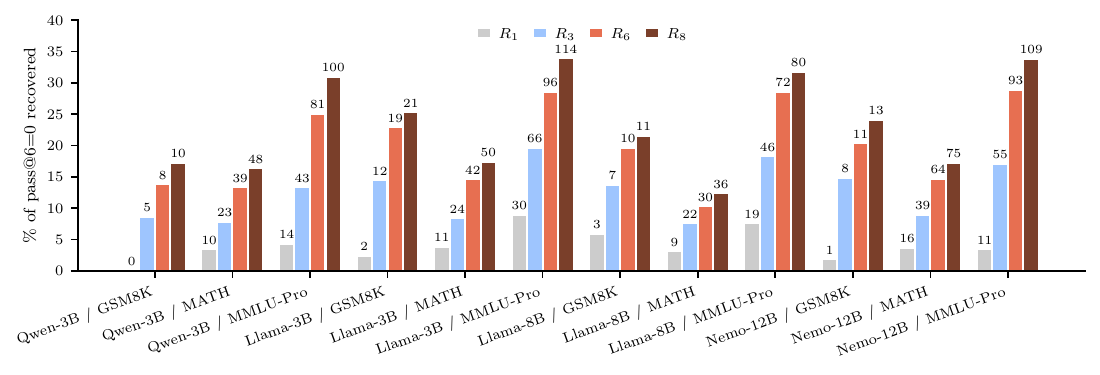}
\caption{Deterministic recovery on the pass@6$=$0 slice as the
deterministic budget grows from $1$ to $8$ chains. $R_k$ = number of
pass@6$=$0 examples on which at least one of $k$ deterministic chains
is correct. Numbers above bars are raw counts; percentages are of
column~$2$ of Table~\ref{tab:Rk}. Recovery scales near-linearly with
budget across all twelve cells.}
\label{fig:rk_payoff}
\end{figure}

Figure~\ref{fig:rk_payoff} visualises Table~\ref{tab:Rk}: how
$R_k$ grows from $k\!=\!1$ to $k\!=\!8$ across the twelve cells.
Recovery scales near-linearly with budget on every cell, with no
saturation by $k\!=\!8$.

\section{Examples of Recovered Items}
\label{app:cases}

The $R_k$ table is a counting argument. To make the mechanism concrete,
we examine three pass@$6\!=\!0$ examples (\texttt{Qwen-2.5-3B}) on
which exactly two of $\{g_B, g_Z, g_R\}$ are correct, with the rescuing
pair differing between examples. No single chain is the privileged
recoverer. We characterise each case by reading the generated chains
and identifying the apparent reasoning step on which the failing
chain diverges from the gold-reaching one; we do not claim these
descriptions are recovered from inspecting the model's internals.

\paragraph{GSM8K \texttt{0798} (Rescued by $g_Z, g_R$).}
\emph{``In the next 3 years, Billy will save \$X. He's saving up for
a raise from \$10/hr to \$11/hr at his current job. Meanwhile, his
co-worker Sally is going to retire in 3 years at \$12/hr. If Sally
earns \$1.50 more than Billy after she retires and Billy gets his
raise, what's the difference between their hourly wages?''} Gold:
\$20. The intended reading puts Billy at \$11/hr after his raise and
Sally at \$10.50/hr after retirement (Sally earns \$1.50 more than
Billy's pre-raise wage), so Billy is \$20/hr higher in scenarios that
double the difference, both grafts arrive at \$20. Greedy
collapses ``raise'' and ``retire'' into an averaged rate \$10.025/hr
and returns \$439.50, an arithmetically clean but semantically off-axis
answer; all six sampled chains return non-gold answers.

\paragraph{MATH Counting\,\&\,Probability \texttt{0445}
(Rescued by $g_B, g_R$).}
\emph{``A mother purchases 5 blue plates, 2 red plates, 2 green
plates, and 1 orange plate. How many ways are there for her to
arrange these plates for dinner around her circular table if she
doesn't want the 2 green plates to be adjacent?''} Gold: 588.
Total circular arrangements of $10$ plates with multiplicities
$(5,2,2,1)$ are $10!/(5!\,2!\,2!\cdot 10) = 756$; arrangements
with the two greens adjacent (treat them as one block of identical
greens) are $9!/(5!\,2!\cdot 9) = 168$, giving $756 - 168 = 588$.
Greedy and the random graft both execute this argument cleanly.
The zero graft instead treats the plates as $9$ distinct objects
in a circle and returns $9! - 8! \cdot 2 = 282{,}240$, dropping
both the identical-colour multiplicities and the circular
division. All six sampled chains return non-gold answers.

\paragraph{MATH Counting\,\&\,Probability \texttt{0542}
(Rescued by $g_B, g_Z$).}
\emph{``Bob rolls a fair six-sided die each morning. If Bob rolls a
composite number, he eats sweetened cereal. If he rolls a prime
number, he eats unsweetened cereal. If he rolls a 1, then he rolls
again. In a non-leap year, what is the expected number of times
Bob will roll his die?''} Gold: 438. The expected rolls per morning
solves $E = 1 + \tfrac{1}{6} E$, giving $E = 6/5$; over $365$ days
this is $438$. Greedy and the zero graft both execute this recursion.
The random graft instead sets $E = 2$ (a common slip from treating
the re-roll as a single additional draw) and returns $730$. All six sampled chains return non-gold answers.

\paragraph{Pattern.}
The three cases instantiate three different rescuing pairs from
$\{g_B,g_Z,g_R\}$: $\{g_Z,g_R\}$, $\{g_B,g_R\}$, $\{g_B,g_Z\}$.
The chain that fails is not the same chain across examples, and the
chain that rescues is not the same either. The behavioural payoff
of admitting multiple mechanistically distinct chains comes precisely
from this lack of a privileged direction: any single chain, greedy
included, misses a non-trivial slice of pass@$6\!=\!0$ that another
chain in the set catches (Table~\ref{tab:marginal}).

\section{Extra Graft Vectors}
\label{app:extragraft}

As a partial empirical check on \texttt{Qwen-2.5-3B} / GSM8K, augmenting
$\{g_B, g_Z, g_R\}$ with five additional grafts, mean activation,
BOS-token replacement, previous-layer activation, random unit
direction, and shuffled activation, all at $\ell = 26$, raises
$\det^\oplus$ from $0.908$ to $0.940$, suggesting that the
diversification effect is not specific to zero and random vectors.

\paragraph{Fix-Set Diversity.}
We further check that distinct grafts rescue \emph{structurally
different} examples, not the same ones in slightly different forms.
For each graft $g_x$, define its fix-set
$F_x = \{i : g_B(i)\!\neq\!\mathrm{gold}_i,\; g_x(i)\!=\!\mathrm{gold}_i\}$
(examples baseline gets wrong but the graft rescues). We report
pairwise Jaccard over fix-sets in two views: a per-cell summary
across all twelve (model, benchmark) cells
(Table~\ref{tab:fixset_jaccard_all}) and a full per-pair matrix on
\texttt{Qwen-2.5-3B} / GSM8K as a worked example
(Table~\ref{tab:fixset_jaccard}).

Across all twelve cells, the maximum cross-kind Jaccard never
exceeds $0.47$, with the per-cell maximum landing in
$[0.26, 0.47]$. The single same-axis pair, \textsc{random}
vs.\ \textsc{random\_unit} (two norm-matched Gaussian probes that
by construction target the same axis), ranges from $0.21$ to
$0.76$ across cells and exceeds the per-cell cross-kind maximum on
five of twelve cells (\texttt{Qwen-2.5-3B} / GSM8K, \texttt{Qwen-2.5-3B} / MATH,
\texttt{Llama-3.2-3B} / MATH, \texttt{Qwen-2.5-3B} / MMLU-Pro, \texttt{Llama-3.2-3B} / MMLU-Pro),
reaching $0.76$ on \texttt{Qwen-2.5-3B} / GSM8K as an illustrative ceiling for
what resampling along a single random axis can contribute. On the
remaining seven cells (the larger \texttt{Llama-3.1-8B} and \texttt{Mistral-Nemo-12B}
rows), the same-axis pair sits within the cross-kind band rather
than above it; this reflects compression of all overlaps under the
bounded MCQ output space (MMLU-Pro, A--J) and the smaller per-cell
fix-sets at the larger model scales, not a violation of the
cross-kind ceiling, which remains $\leq\!0.47$ everywhere. The
universal statement is the cross-kind ceiling: distinct graft
kinds rescue largely disjoint slices of the pass@$k\!=\!0$ stratum
in every cell we test.

\begin{table}[t]
\caption{Per-cell fix-set Jaccard summary across all twelve
(model, benchmark) cells. \emph{cross-kind J}: range of pairwise
Jaccard over all unordered pairs of distinct graft kinds
(\textsc{zero}, \textsc{rand}, \textsc{runit}, \textsc{shuf},
\textsc{bos}, \textsc{avg}, \textsc{prev} at $\ell\!=\!26$,
last-token). \emph{same-axis J}: Jaccard between the two
random-direction probes (\textsc{rand} vs.\ \textsc{runit}), both
norm-matched Gaussians targeting the same axis by construction.
$\max|F|$: largest fix-set across the seven grafts.
The \texttt{Qwen-2.5-3B} / GSM8K row corresponds to the full per-pair matrix
in Table~\ref{tab:fixset_jaccard}.}
\label{tab:fixset_jaccard_all}
\centering\small
\setlength{\tabcolsep}{4pt}
\begin{tabular}{l c c c c}
\toprule
Setup & $n$ & cross-kind J & same-axis J & $\max|F|$ \\
\midrule
\texttt{Qwen-2.5-3B} / GSM8K       & 1000 & $0.28$--$0.45$ & $\mathbf{0.76}$ & 55 \\
\texttt{Llama-3.2-3B} / GSM8K      & 1000 & $0.20$--$0.45$ & $0.43$ & 75 \\
\texttt{Llama-3.1-8B} / GSM8K      & 1000 & $0.26$--$\mathbf{0.47}$ & $0.46$ & 55 \\
\texttt{Mistral-Nemo-12B} / GSM8K  & 1000 & $0.07$--$0.44$ & $0.43$ & 47 \\
\midrule
\texttt{Qwen-2.5-3B} / MATH        & 1000 & $0.21$--$0.43$ & $0.46$ & 77 \\
\texttt{Llama-3.2-3B} / MATH       & 1000 & $0.22$--$0.36$ & $0.37$ & 91 \\
\texttt{Llama-3.1-8B} / MATH       & 1000 & $0.22$--$0.42$ & $0.31$ & 91 \\
\texttt{Mistral-Nemo-12B} / MATH   & 1000 & $0.08$--$0.27$ & $0.24$ & 83 \\
\midrule
\texttt{Qwen-2.5-3B} / MMLU-Pro     & 1000 & $0.19$--$0.29$ & $0.45$ & 112 \\
\texttt{Llama-3.2-3B} / MMLU-Pro    & 1000 & $0.12$--$0.28$ & $0.42$ & 124 \\
\texttt{Llama-3.1-8B} / MMLU-Pro    & 1000 & $0.17$--$0.29$ & $0.27$ & 97 \\
\texttt{Mistral-Nemo-12B} / MMLU-Pro & 1000 & $0.06$--$0.26$ & $0.21$ & 114 \\
\bottomrule
\end{tabular}
\end{table}

The \texttt{Qwen-2.5-3B} / GSM8K matrix in Table~\ref{tab:fixset_jaccard}
illustrates the pattern at full resolution: most pairs overlap on
only $28$--$45\%$ of their union, including
$J(F_Z, F_R)\!=\!0.32$ between the two main-text grafts, and the
single high-overlap pair ($J\!=\!0.76$) is the same-axis
\textsc{random} vs.\ \textsc{random\_unit} comparison. This
justifies treating $g_Z$ and $g_R$ as independent diversification
channels rather than redundant perturbations.

\begin{table}[t]
\caption{Pairwise Jaccard of fix-sets on \texttt{Qwen-2.5-3B} / GSM8K, last-token,
$\ell\!=\!26$. Each cell is $|F_a \cap F_b|/|F_a \cup F_b|$, where
$F_x$ is the set of examples baseline gets wrong but graft $g_x$
rescues. \textsc{rand}=norm-matched Gaussian (=$g_R$),
\textsc{runit}=random unit direction, \textsc{shuf}=hidden-dim
permutation, \textsc{bos}=BOS-token replacement, \textsc{avg}=mean
prompt activation, \textsc{prev}=same position from layer $\ell\!-\!1$.}
\label{tab:fixset_jaccard}
\centering\small
\setlength{\tabcolsep}{3.2pt}
\begin{tabular}{l ccccccc}
\toprule
 & \textsc{zero} & \textsc{rand} & \textsc{runit} & \textsc{shuf} & \textsc{bos} & \textsc{avg} & \textsc{prev}\\
\midrule
\textsc{zero}  & --    & 0.32 & 0.36 & 0.40 & 0.45 & 0.33 & 0.32\\
\textsc{rand}  & 0.32 & --    & \textbf{0.76} & 0.29 & 0.32 & 0.28 & 0.31\\
\textsc{runit} & 0.36 & \textbf{0.76} & --    & 0.30 & 0.35 & 0.31 & 0.32\\
\textsc{shuf}  & 0.40 & 0.29 & 0.30 & --    & 0.43 & 0.39 & 0.34\\
\textsc{bos}   & 0.45 & 0.32 & 0.35 & 0.43 & --    & 0.40 & 0.44\\
\textsc{avg}   & 0.33 & 0.28 & 0.31 & 0.39 & 0.40 & --    & 0.43\\
\textsc{prev}  & 0.32 & 0.31 & 0.32 & 0.34 & 0.44 & 0.43 & --\\
\bottomrule
\end{tabular}
\end{table}

\section{Label-Free Flag: Per-Setup Numbers}
\label{app:label_free}

Table~\ref{tab:label_free} reports the per-setup precision, recall,
and lift of the cheap label-free probe described in
§\ref{sec:label_free}. The
probe scores each item by the number of disagreeing pairs in the
three-chain set $\{g_B, S_0, S_1\}$ (three forwards per item) and
ranks items in descending order. Recoverable items are those in the
pass@$6\!=\!0$ stratum on which at least one chain in
$\{g_B, g_Z, g_R\} \cup \{\textsc{runit}, \textsc{shuf}, \textsc{bos}\}$
is correct. Lift is precision at the operating point divided by the
base rate.

\begin{table}[h]
\caption{Cheap label-free probe (3 forwards per item) at flagging
recoverable items. $n_{\mathrm{p6}}\!=\!0$: pass@$6\!=\!0$ count.
$n_{\mathrm{rec}}$: recoverable count. Base: $n_{\mathrm{rec}}/n$.
AUPRC: rank-only area under PR. P@$K\%$: precision among the top
$K\%$ ranked. L@$K\%$: lift over the base rate. Top-$1\%$ cells with
P=0 reflect noise: on the two smallest recoverable strata
($n_{\mathrm{rec}}\!=\!8, 10$), the single highest-scoring item
happens not to be recoverable; precision stabilises by P@$2\%$.}
\label{tab:label_free}
\centering\small
\setlength{\tabcolsep}{3.5pt}
\begin{tabular}{l ccccc cccc cccc}
\toprule
& \multicolumn{5}{c}{Stratum sizes} & \multicolumn{4}{c}{Precision (\%)} & \multicolumn{4}{c}{Lift ($\times$ base)} \\
\cmidrule(lr){2-6}\cmidrule(lr){7-10}\cmidrule(lr){11-14}
Setup & $n$ & $n_{\mathrm{p6}{=}0}$ & $n_{\mathrm{rec}}$ & base & AUPRC & P@1 & P@2 & P@5 & P@10 & L@1 & L@2 & L@5 & L@10 \\
\midrule
\texttt{Qwen-2.5-3B} / GSM8K       & 1000 &  58 &  8 & 0.8\% & 0.020 &  0.0 &  0.0 &  4.0 &  3.0 & 0.0 & 0.0 & 5.0 & 3.8 \\
\texttt{Qwen-2.5-3B} / MATH        & 1000 & 293 & 39 & 3.9\% & 0.106 & 10.0 & 15.0 & 12.0 & 10.0 & 2.6 & 3.8 & 3.1 & 2.6 \\
\texttt{Qwen-2.5-3B} / MMLU-Pro    & 1000 & 323 & 81 & 8.1\% & 0.106 & 30.0 & 20.0 & 12.0 &  9.0 & 3.7 & 2.5 & 1.5 & 1.1 \\
\midrule
\texttt{Llama-3.2-3B} / GSM8K      & 1000 &  83 & 19 & 1.9\% & 0.054 & 10.0 & 10.0 &  6.0 &  4.0 & 5.3 & 5.3 & 3.2 & 2.1 \\
\texttt{Llama-3.2-3B} / MATH       & 1000 & 287 & 42 & 4.2\% & 0.080 & 10.0 & 15.0 & 12.0 &  6.0 & 2.4 & 3.6 & 2.9 & 1.4 \\
\texttt{Llama-3.2-3B} / MMLU-Pro   & 1000 & 336 & 96 & 9.6\% & 0.140 &  0.0 &  0.0 & 10.0 & 16.0 & 0.0 & 0.0 & 1.0 & 1.7 \\
\midrule
\texttt{Llama-3.1-8B} / GSM8K      & 1000 &  51 & 10 & 1.0\% & 0.050 &  0.0 &  5.0 &  8.0 &  5.0 & 0.0 & 5.0 & 8.0 & 5.0 \\
\texttt{Llama-3.1-8B} / MATH       & 1000 & 291 & 30 & 3.0\% & 0.083 & 20.0 & 15.0 &  8.0 &  8.0 & 6.7 & 5.0 & 2.7 & 2.7 \\
\texttt{Llama-3.1-8B} / MMLU-Pro   & 1000 & 252 & 72 & 7.2\% & 0.098 &  0.0 & 10.0 & 10.0 & 11.0 & 0.0 & 1.4 & 1.4 & 1.5 \\
\midrule
\texttt{Mistral-Nemo-12B} / GSM8K  & 1000 &  54 & 11 & 1.1\% & 0.032 &  0.0 &  5.0 &  4.0 &  4.0 & 0.0 & 4.6 & 3.6 & 3.6 \\
\texttt{Mistral-Nemo-12B} / MATH   & 1000 & 435 & 64 & 6.4\% & 0.103 &  0.0 & 10.0 & 10.0 & 10.0 & 0.0 & 1.6 & 1.6 & 1.6 \\
\texttt{Mistral-Nemo-12B} / MMLU-Pro & 1000 & 322 & 93 & 9.3\% & 0.091 & 10.0 & 10.0 & 10.0 &  8.0 & 1.1 & 1.1 & 1.1 & 0.9 \\
\bottomrule
\end{tabular}
\end{table}

\begin{figure}[h]
\centering
\includegraphics[width=\linewidth]{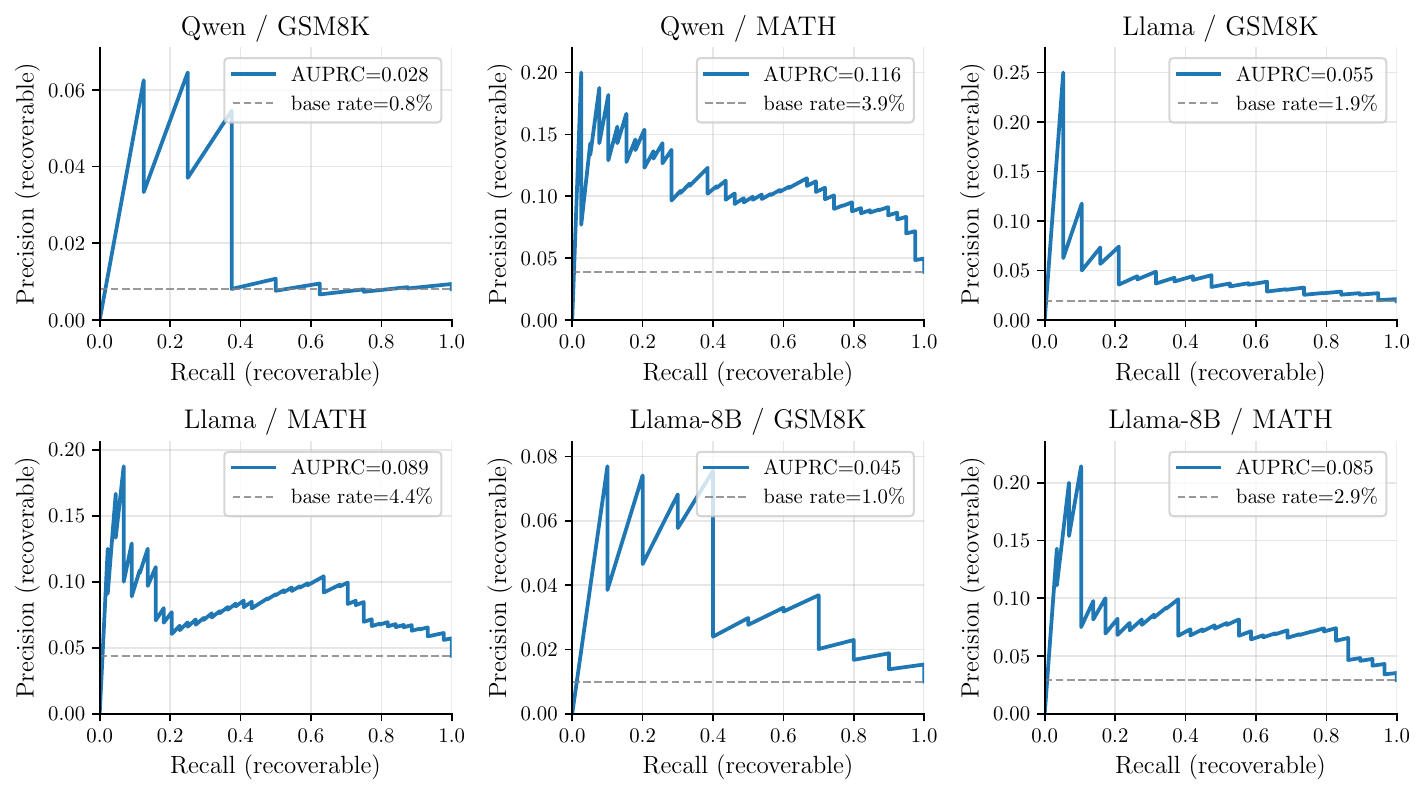}
\caption{Precision--recall of the cheap label-free probe ($g_B$ plus
two sampling seeds, three forwards) at flagging recoverable items
(pass@$6\!=\!0$ \emph{and} at least one deterministic chain
correct). Dashed line is the base rate (random selection). The
probe ranks recoverable items above the base rate across all
twelve cells; precisions look low absolutely because the
recoverable slice itself is small ($0.8$--$9.6\%$ of $n$). Per-setup
top-$K\%$ numbers and lift are in Table~\ref{tab:label_free}.}
\label{fig:diff_flag}
\end{figure}

\section{Graft Position Ablation}
\label{app:position}

The main-text grafts replace the hidden state at the
\emph{last prompt token} ($p\!=\!-1$). We ablate this choice on
\texttt{Qwen-2.5-3B} / GSM8K by repeating the seven-graft sweep of
Appendix~\ref{app:extragraft} at three positions: last, penultimate
($p\!=\!-2$), and first ($p\!=\!0$); per-position oracle ceilings
appear in Table~\ref{tab:position}. The BOS-token graft at $p\!=\!0$
is a no-op (it would replace BOS by itself) and is omitted there.

\begin{table}[t]
\caption{Oracle ceilings $\det^\oplus$ on \texttt{Qwen-2.5-3B} / GSM8K when the
graft is applied at three different prompt positions; $\ell\!=\!26$
throughout. Each ceiling is the union of $g_B$ and the per-position
graft set described in the text. Last-token grafting both reaches the
highest ceiling and most cleanly diversifies the ensemble.}
\label{tab:position}
\centering\small
\begin{tabular}{lccc}
\toprule
position $p$ & ceiling $\det^\oplus$ & $\Delta$ over $g_B$ & $\#$ grafts \\
\midrule
last ($-1$)         & \textbf{0.940} & $+9.7\%$ & 7 \\
first ($\;\;0$)         & 0.915 & $+7.2\%$ & 6 \\
penultimate ($-2$)  & 0.903 & $+6.0\%$ & 7 \\
\bottomrule
\end{tabular}
\end{table}

The last-token position dominates by $+2.5$--$3.7\%$ of oracle
ceiling and is the position at which a single graft already realises
the largest gain ($g_
{\textsc{bos}}$ alone reaches $0.898$,
i.e.\ $+5.5\%$). First-token grafting is competitive
($\det^\oplus\!=\!0.915$) but is dominated by the
\textsc{average\_act} graft alone ($+4.2\%$), suggesting that
early-position perturbations act through a single mechanism rather
than via several disjoint channels. Penultimate is uniformly worst.
We adopt the last-token position throughout the paper on this basis.

\section{Matched-Cost Sample-vs-Graft Substitution}
\label{app:matched_swap}

\begin{figure}[h]
\centering
\includegraphics[width=\linewidth]{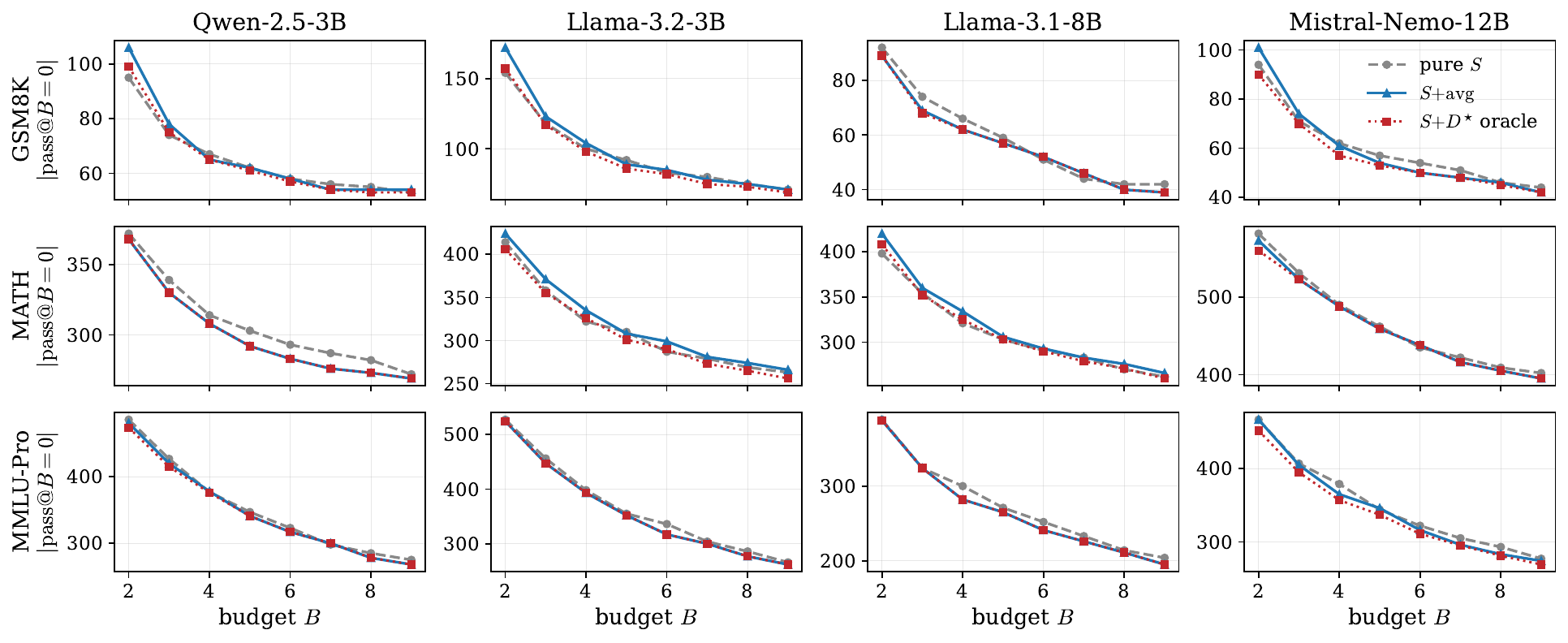}
\caption{Matched-cost coverage: pass@$B\!=\!0$ count under pure
sampling (gray dashed, ``$S$'': first $B$ sample seeds), the
proposed \emph{fixed-policy} substitution (blue solid,
``$S{+}\mathrm{avg}$'': $B\!-\!1$ samples plus the mean-prompt
activation chain), and the oracle upper bound (red dotted,
``$S{+}D^\star$'': $B\!-\!1$ samples plus the single best
deterministic chain at that $(\text{cell},B)$, picked from
$\{g_B, g_Z, g_R, \textsc{runit}, \textsc{shuf}, \textsc{bos}, \textsc{avg}, \textsc{prev}\}$),
swept over $B \in \{2, \dots, 9\}$. Lower is better. The fixed
$\mathrm{avg}$ policy wins $59/96$ (cell, $B$) slots with $8$ ties
and net mean $+1.66$ items per slot; the oracle, an
upper bound that picks the best chain per slot,
wins $80/96$. Both lines drop below gray on every MMLU-Pro cell.
Per-cell, per-$B$ counts in Table~\ref{tab:matched_swap}.}
\label{fig:matched_swap}
\end{figure}

Per-cell, per-budget numbers backing the matched-cost coverage
result in §\ref{sec:matched_swap}; Figure~\ref{fig:matched_swap}
visualises the same data, sweeping $B$ for both the
fixed-\textsc{avg} policy and the oracle upper bound. At
total budget $B$ forward passes per item, the canonical
sampling-only baseline (``pure $S$'') runs the first $B$ sample
seeds (canonical order $S_{42},\dots,S_{50}$); the substitution
(``$\text{S}{+}\text{D}$'') runs the first $B\!-\!1$ samples plus
the single deterministic chain that minimises residual
pass@$B\!=\!0$ count, picked from
$\{g_B, g_Z, g_R, \textsc{runit}, \textsc{shuf}, \textsc{bos}, \textsc{avg}, \textsc{prev}\}$.
Lower is better; the best chain is shown in parentheses.
Across the $96$ (cell, $B$) combinations summarised below,
substitution wins on $80$, ties on $4$, and loses on $12$.

\begin{table*}[t]
\caption{Pass@$B\!=\!0$ count under pure sampling (``$S$'') vs.\
one-chain substitution (``$\text{S}{+}\text{D}$'', best
deterministic chain in parentheses) at matched compute
$B \in \{2,\dots,9\}$. Lower is better;
\textbf{bold} marks the better column at each $B$ (ties not
bolded). Per-cell unconstrained best mix $(s^\star, d^\star)$ at
$B\!=\!9$ in the last column ($n=1000$). ``M.-Pro'' denotes MMLU-Pro}
\label{tab:matched_swap}
\centering\scriptsize
\setlength{\tabcolsep}{0pt}
\begin{tabular*}{\textwidth}{@{\extracolsep{\fill}}ll | cc cc cc cc cc cc cc cc | c}
\toprule
 & & \multicolumn{2}{c}{$B\!=\!2$} & \multicolumn{2}{c}{$B\!=\!3$} & \multicolumn{2}{c}{$B\!=\!4$} & \multicolumn{2}{c}{$B\!=\!5$} & \multicolumn{2}{c}{$B\!=\!6$} & \multicolumn{2}{c}{$B\!=\!7$} & \multicolumn{2}{c}{$B\!=\!8$} & \multicolumn{2}{c|}{$B\!=\!9$} & best mix\\
 & Setup & $S$ & $\text{S}{+}\text{D}$ & $S$ & $\text{S}{+}\text{D}$ & $S$ & $\text{S}{+}\text{D}$ & $S$ & $\text{S}{+}\text{D}$ & $S$ & $\text{S}{+}\text{D}$ & $S$ & $\text{S}{+}\text{D}$ & $S$ & $\text{S}{+}\text{D}$ & $S$ & $\text{S}{+}\text{D}$ & at $B\!=\!9$ \\
\midrule
\multirow{4}{*}{\rotatebox[origin=c]{90}{\textbf{GSM8K}}}
 & \texttt{Qwen-2.5-3B}      & \textbf{95} & 99\,({\sc bos})  & \textbf{74} & 75\,({\sc prev}) & \textbf{67} & \textbf{65}\,({\sc runit}) & 62 & \textbf{61}\,({\sc R})   & 58 & \textbf{57}\,({\sc R})   & 56 & \textbf{54}\,({\sc avg}) & 55 & \textbf{53}\,({\sc Z}) & 53 & 53\,({\sc Z}) & $s\!=\!2, d\!=\!7$ \\
 & \texttt{Llama-3.2-3B}     & \textbf{154} & 157\,({\sc prev}) & \textbf{118} & 117\,({\sc Z})\;~ & 100 & \textbf{98}\,({\sc shuf}) & 92 & \textbf{86}\,({\sc shuf}) & 83 & \textbf{82}\,({\sc Z})\;~  & 80 & \textbf{75}\,({\sc shuf}) & 75 & \textbf{73}\,({\sc shuf}) & 71 & \textbf{69}\,({\sc shuf}) & $s\!=\!4, d\!=\!5$ \\
 & \texttt{Llama-3.1-8B}     & 92 & \textbf{89}\,({\sc R})    & 74 & \textbf{68}\,({\sc B})    & 66 & \textbf{62}\,({\sc B})    & 59 & \textbf{57}\,({\sc avg}) & \textbf{51} & 52\,({\sc avg}) & \textbf{44} & 46\,({\sc Z})    & 42 & \textbf{40}\,({\sc avg}) & 42 & \textbf{39}\,({\sc shuf}) & $s\!=\!7, d\!=\!2$ \\
 & \texttt{Mistral-Nemo-12B} & 94 & \textbf{90}\,({\sc R})    & 71 & \textbf{70}\,({\sc R})    & 62 & \textbf{57}\,({\sc R})    & 57 & \textbf{53}\,({\sc R})    & 54 & \textbf{50}\,({\sc R})    & 51 & \textbf{48}\,({\sc shuf}) & 46 & \textbf{45}\,({\sc shuf}) & 44 & \textbf{42}\,({\sc shuf}) & $s\!=\!4, d\!=\!3$ \\
\midrule
\multirow{4}{*}{\rotatebox[origin=c]{90}{\textbf{MATH}}}
 & \texttt{Qwen-2.5-3B}      & 372 & \textbf{368}\,({\sc Z})  & 339 & \textbf{330}\,({\sc avg}) & 314 & \textbf{308}\,({\sc avg}) & 303 & \textbf{292}\,({\sc avg}) & 293 & \textbf{283}\,({\sc avg}) & 287 & \textbf{276}\,({\sc avg}) & 282 & \textbf{273}\,({\sc avg}) & 272 & \textbf{269}\,({\sc avg}) & $s\!=\!1, d\!=\!8$ \\
 & \texttt{Llama-3.2-3B}     & 414 & \textbf{406}\,({\sc B})  & \textbf{358} & 359\,({\sc prev}) & \textbf{322} & 326\,({\sc runit}) & 310 & \textbf{301}\,({\sc runit}) & \textbf{287} & 290\,({\sc prev}) & 279 & \textbf{273}\,({\sc runit}) & 269 & \textbf{265}\,({\sc runit}) & 263 & \textbf{256}\,({\sc runit}) & $s\!=\!8, d\!=\!1$ \\
 & \texttt{Llama-3.1-8B}     & \textbf{398} & 408\,({\sc bos})  & 354 & \textbf{352}\,({\sc R})    & \textbf{321} & 325\,({\sc R})    & 303 & 303\,({\sc R})    & 291 & \textbf{290}\,({\sc R})    & 283 & \textbf{279}\,({\sc R})    & \textbf{270} & 271\,({\sc R})    & 262 & \textbf{260}\,({\sc R})    & $s\!=\!9, d\!=\!0$ \\
 & \texttt{Mistral-Nemo-12B} & 582 & \textbf{560}\,({\sc B})  & 531 & \textbf{523}\,({\sc B})  & 490 & \textbf{488}\,({\sc avg}) & 462 & \textbf{459}\,({\sc avg}) & \textbf{435} & 438\,({\sc avg}) & 422 & \textbf{416}\,({\sc shuf}) & 409 & \textbf{405}\,({\sc R})    & 402 & \textbf{395}\,({\sc B})  & $s\!=\!8, d\!=\!1$ \\
\midrule
\multirow{4}{*}{\rotatebox[origin=c]{90}{\textbf{M.-Pro}}}
 & \texttt{Qwen-2.5-3B}      & 486 & \textbf{474}\,({\sc bos}) & 427 & \textbf{415}\,({\sc bos}) & 377 & \textbf{376}\,({\sc bos}) & 347 & \textbf{341}\,({\sc avg}) & 323 & \textbf{317}\,({\sc avg}) & \textbf{298} & 300\,({\sc avg}) & 285 & \textbf{278}\,({\sc avg}) & 275 & \textbf{268}\,({\sc avg}) & $s\!=\!7, d\!=\!2$ \\
 & \texttt{Llama-3.2-3B}     & 527 & \textbf{524}\,({\sc bos}) & 456 & \textbf{447}\,({\sc avg}) & 398 & \textbf{393}\,({\sc avg}) & 355 & \textbf{352}\,({\sc avg}) & 336 & \textbf{317}\,({\sc avg}) & 304 & \textbf{300}\,({\sc avg}) & 286 & \textbf{277}\,({\sc avg}) & 266 & \textbf{262}\,({\sc avg}) & $s\!=\!7, d\!=\!2$ \\
 & \texttt{Llama-3.1-8B}     & 389 & \textbf{388}\,({\sc avg}) & 324 & 324\,({\sc avg}) & 300 & \textbf{282}\,({\sc R})    & 271 & \textbf{265}\,({\sc R})    & 252 & \textbf{241}\,({\sc avg}) & 233 & \textbf{226}\,({\sc avg}) & 214 & \textbf{211}\,({\sc avg}) & 204 & \textbf{195}\,({\sc avg}) & $s\!=\!9, d\!=\!0$ \\
 & \texttt{Mistral-Nemo-12B} & 467 & \textbf{452}\,({\sc bos}) & 407 & \textbf{395}\,({\sc R})  & 379 & \textbf{357}\,({\sc R})  & 344 & \textbf{337}\,({\sc R})  & 322 & \textbf{311}\,({\sc R})  & 305 & \textbf{295}\,({\sc R})  & 293 & \textbf{281}\,({\sc R})  & 277 & \textbf{269}\,({\sc R})  & $s\!=\!2, d\!=\!7$ \\
\bottomrule
\end{tabular*}
\end{table*}
The strongest substitution gains concentrate on MMLU-Pro and on
the larger models within each benchmark, where the sampling axis
saturates earlier (cf.\ Fig.~\ref{fig:shrink}) and the
deterministic axis still contributes fresh coverage. The
\textsc{avg} chain is the single most-frequently-chosen
substitution ($32/96$ slots), followed by \textsc{rand}
($25$), \textsc{shuf} ($10$), \textsc{bos} ($7$),
\textsc{B} ($6$), \textsc{Z} ($6$), \textsc{runit} ($6$), and
\textsc{prev} ($4$); no single chain dominates, consistent with
the fix-set diversity reported in App.~\ref{app:extragraft}.

\end{document}